%% file: main.tex
\documentclass[11pt]{article}

\usepackage[preprint]{acl}

\usepackage{times}
\usepackage{latexsym}
\usepackage{comment}
\usepackage[T1]{fontenc}

\usepackage[utf8]{inputenc}

\usepackage{microtype}

\usepackage{inconsolata}

\usepackage{graphicx}
\usepackage{adjustbox}
\usepackage{multirow}
\usepackage{booktabs}        
\usepackage{amsmath} 
\usepackage{makecell}
\usepackage{listings}
\title{Business Logic-Driven Text-to-SQL Data Synthesis for Business Intelligence}

\author{
Jinhui Liu$^{1}$,
Ximeng Zhang$^{2}$,
Yanbo Ai$^{2}$,
Zhou Yu$^{1}$\thanks{Corresponding author.} \\
$^{1}$Department of Computer Science, Columbia University, 
$^{2}$Mercury \\
\texttt{\{jl7309, zy2461\}@columbia.edu}
}

\begin{document}
\maketitle
\begin{abstract}
Evaluating Text-to-SQL agents in private business intelligence (BI) settings is challenging due to the scarcity of realistic, domain-specific data. 
While synthetic evaluation data offers a scalable solution, existing generation methods fail to capture business realism--whether questions reflect realistic business logic and workflows. 
We propose a \textbf{Business Logic–Driven Data Synthesis} framework that generates data grounded in business personas, work scenarios, and workflows. In addition, we improve the data quality by imposing a \textbf{business reasoning complexity control} strategy that diversifies the analytical reasoning steps required to answer the questions. 
Experiments on a production-scale Salesforce database show that our synthesized data achieves high business realism (\textbf{98.44\%}), substantially outperforming OmniSQL (+19.5\%) and SQL-Factory (+54.7\%), while maintaining strong question–SQL alignment (\textbf{98.59\%}). 
Our synthetic data also reveals that state-of-the-art Text-to-SQL models still have significant performance gaps, achieving only 42.86\% execution accuracy on the most complex business queries.
\end{abstract}

\begin{figure*}[t]
\centering
  \includegraphics[width=\linewidth]{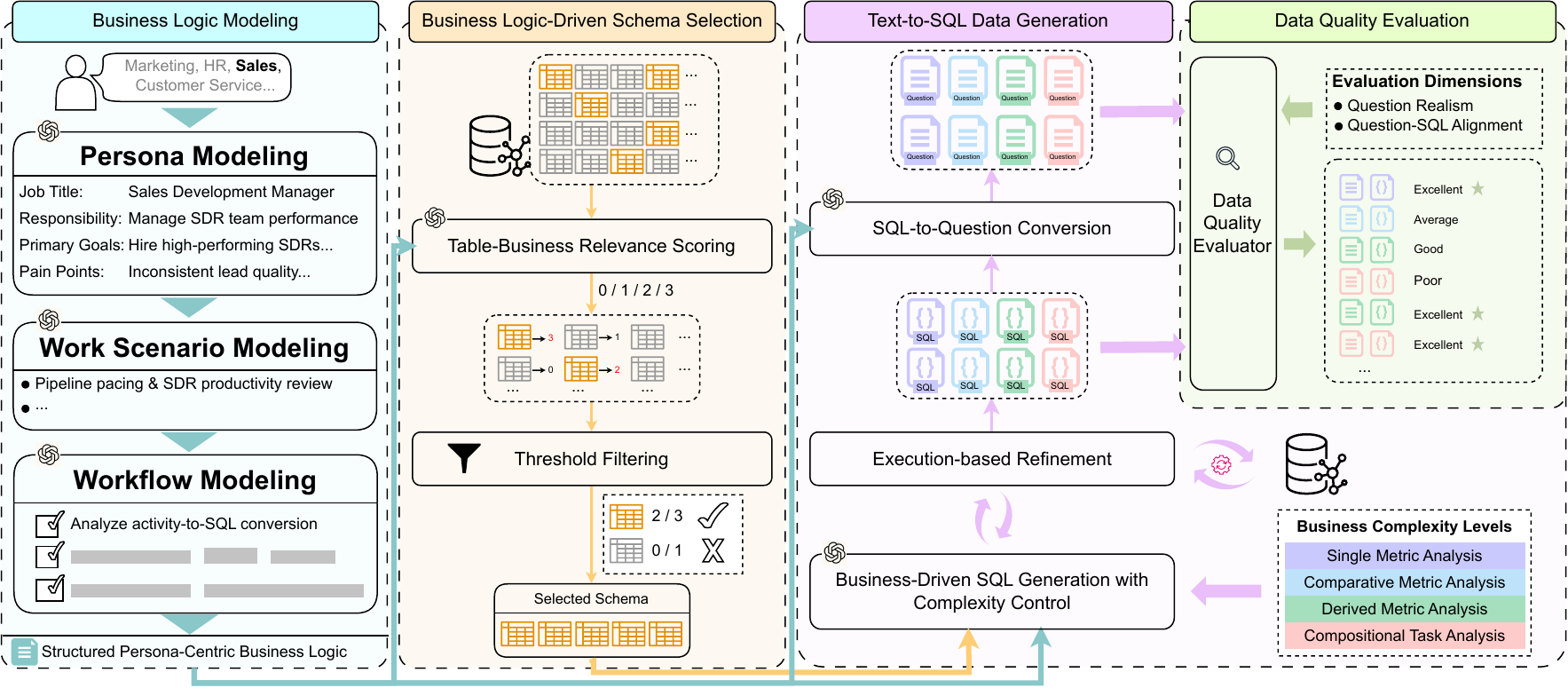}
  \caption {Overview of the proposed \textbf{Business Logic–Driven Text-to-SQL Data Synthesis framework}.
  The framework starts with \textbf{business logic modeling} (see left), where business logic is modeled as personas, work scenarios, and workflows. This structured business logic guides \textbf{business logic–driven schema selection} (see middle), which scores and filters table schema based on business logic relevance. Conditioned on the selected schema and business logic context, the framework performs \textbf{Text-to-SQL Data Generation} (see right), comprising \textbf{business-driven SQL generation with reasoning complexity control} with execution-based refinement and \textbf{SQL-to-question conversion}, to synthesize executable queries across diverse business reasoning complexity levels and business-oriented natural language questions. Finally, a \textbf{data quality evaluation module} assesses question realism and question–SQL alignment, ensuring high-quality, business-realistic Text-to-SQL evaluation data (see top right).}
  \label{overview}
\end{figure*}

\section{Introduction}
Text-to-SQL enables natural language access to structured data and is a core capability in business intelligence (BI), where non-technical users (e.g., analysts, managers, and executives) rely on it to explore enterprise data and support decision making\cite{SurveyTexttoSQLEra2025,ImpactBI}.
Recent advances in large language models (LLMs) have substantially improved Text-to-SQL performance and accelerated the deployment of LLM-based Text-to-SQL agents in BI systems, increasing the need for reliable evaluation of the agents on private enterprise databases.

Constructing evaluation datasets in private BI settings is challenging due to the cost, scalability, and privacy constraints of manual annotation.
Therefore, synthetic data generation has been a practical alternative. 
However, existing data generation methods prioritize broad schema coverage, diverse SQL patterns, and syntactic complexity but pay limited attention to business realism\cite{OmniSQL,SQL-Factory,LearnedSQLGen}.
As a result, the generated questions fail to capture how questions arise in real BI workflows and are detached from realistic business logic. This mismatch limits the ability of existing synthetic datasets to faithfully evaluate how Text-to-SQL agents perform in realistic enterprise scenarios.

In real BI scenarios, questions do not arise independently from the database schema. 
Instead, they are driven by business logic that reflects how people work in a business domain. Each question is asked by a specific job role working in a particular work scenario and is motivated by a concrete task from the corresponding workflow in the scenario \cite{CRMArena}. 
For example, a sales manager reviewing weekly pipeline health may query the database to identify underperforming regions, whereas an executive preparing a quarterly forecast may query the same data to assess overall revenue trends. 
Therefore, even when querying the same set of tables, people with different job roles, under different work scenarios and workflows,  would ask different types of questions.
 
Inspired by how questions arise in real BI workflows, we propose a \textbf{Business Logic-Driven Data Synthesis} framework to synthesize business-realistic Text-to-SQL evaluation data in private BI settings.
Our framework explicitly models business logic as a hierarchical structure of (i) \textbf{personas} representing realistic job roles and responsibilities, (ii) \textbf{work scenarios} capturing recurring business situations for each persona, and (iii) \textbf{workflows} that decompose each scenario into concrete analytical tasks. This structured business logic is used to guide schema selection, SQL generation, and question formulation, enabling the synthesis of question–SQL pairs that reflect how data is queried and used in real-world business scenarios.

Existing data generation approaches characterize query difficulty using SQL syntactic features, such as the number of joins or nesting depth \cite{SQL-Factory,SQLBarber}. However, these features are poorly aligned with business complexity. 
Many business questions that involve simple SQL structures can require complex reasoning (e.g., apply domain-specific rules, or filter objects according to implicit business constraints). 

To better reflect business complexity, we introduce a \textbf{business reasoning complexity control} strategy that characterizes query difficulty by the reasoning steps and patterns required to answer a business question. 
With the business reasoning complexity control, we systematically generate data with diverse reasoning steps and patterns that reflect real-world BI analysis.

To evaluate the data quality, we apply LLM-as-a-Judge to assess \textbf{question realism} and \textbf{question–SQL alignment}, which measure the business realism and semantic correctness of synthesized data. 
We conduct experiments on a production-scale Salesforce database and find that our framework substantially improves business realism, with our dataset achieving 98.44\%, outperforming OmniSQL by +19.5\% and SQL-Factory by +53.7\%, while maintaining strong question–SQL alignment (98.59\%). We further use the data for Text-to-SQL agent evaluation and observe that even the strongest model achieves only 42.86\% execution accuracy on the most complex questions. 
These results indicate that the dataset is both realistic and challenging.

In summary, our main contributions are:
\begin{itemize}
  \item We propose a \textbf{Business Logic–Driven Data Synthesis} framework that synthesizes business realistic Text-to-SQL evaluation data in private BI settings by modeling business logic as a hierarchy of personas, work scenarios, and workflows.
  \item We introduce a \textbf{Business Reasoning Complexity Control} strategy that defines query difficulty by business reasoning steps and patterns, enabling systematic control over the analytical difficulty of generated data.
  \item We present an empirical study on a production-scale enterprise database showing that the synthesized dataset achieves high business realism and semantic correctness, and reveals substantial performance gaps in state-of-the-art Text-to-SQL agents.
\end{itemize}

\section{Related Work}
\input{latex/section/related_work}

\section{Method}
We propose a modular framework for synthesizing Text-to-SQL evaluation datasets that reflect realistic business logic on private enterprise databases. The framework grounds data generation in structured business logic and controls the business reasoning complexity of each query.

As illustrated in Figure \ref{overview}, our framework consists of three main modules:
(1) Business Logic Modeling (Section~\ref{method_1}) which models business logic with personas, work scenarios, and workflows; (2) Business Logic-Driven Schema Selection (Section \ref{method_2}) which select database schema subset relevant to business logic, and (3) Text-to-SQL Data Generation (Section \ref{method_3}), which comprises two submodules: (i) Business-Driven SQL Generation with Business Reasoning Complexity Control,  which generate business realistic SQL queries with different business complexity levels, and (ii) SQL-to-question conversion, which convert SQL queries into business-oriented natural language question.
We also propose an LLM-as-a-Judge data quality evaluation module in (4) Data Quality Evaluation Metrics (Section \ref{method_5}).

\subsection{Business Logic Modeling}\label{method_1}
To improve business realism in synthetic Text-to-SQL data, we explicitly model business logic and incorporate it into the data generation pipeline.
Following how analytical questions arise in real BI settings, we represent business logic as a hierarchical structure with three components: \textbf{persona}, \textbf{work scenario}, and \textbf{workflow}.

A \textbf{persona} represents a realistic job role within a business domain (e.g., sales manager, operations analyst), characterized by its responsibilities, primary goals, and pain points.
A work \textbf{scenario} captures a business situation in which a persona engages with data to make decisions, or monitor outcomes to fulfill their responsibilities (e.g., pipeline monitoring, performance review). The same persona may ask different questions under different scenarios, even over the same database.
Each work scenario is further decomposed into a \textbf{workflow}, modeled as a sequence of analytical tasks that the persona performs to complete the work. These tasks (e.g., filtering data, comparing metrics) naturally give rise to specific analytical questions.

Specifically, business logic generation proceeds in a hierarchical manner (see left column in Figure \ref{overview}). Given a target business domain, we generate a diverse set of personas that cover different departments and seniority levels. For each persona, we generate multiple work scenarios corresponding to different responsibility areas. We then decompose each work scenario into a detailed workflow that consists of multiple analytical tasks. 
This hierarchical modeling (persona → scenario → workflow) yields a collection of structured persona-centric business logic that reflects real business activities.

With the structured business logic, the data generation process is guided by concrete work objectives rather than the database schema alone, resulting in more business-realistic Text-to-SQL data.

\subsection{Business Logic-Driven Schema Selection}\label{method_2}
Production-scale business databases often contain hundreds or thousands of tables, making it impractical to expose the full database schema to the data generation process. In real BI scenarios, analysts reason about a small subset of relevant tables based on their task and intent. We mimic this behavior and propose a business logic-driven schema selection mechanism (see middle column in Figure \ref{overview}).

Given a structured business logic instance, we score database tables according to their relevance to the business logic context, using four discrete relevance levels (0–3). Tables that directly represent core business entities or key metrics receive higher relevance (2-3), while tables that provide auxiliary or unrelated information receive lower scores (0-1). Only the higher-relevance tables are retained for data generation. This step reduces schema complexity while preserving the tables necessary to generate realistic business queries.

\subsection{Text-to-SQL Data Generation}
\paragraph{Business-Driven SQL Generation with Business Reasoning Complexity Control}\label{method_3}

We generate SQL queries by grounding the data generation process in structured business logic and a selected schema subset.
We adopt the SQL-first generation strategy, treating the SQL query as the executable ground truth from which natural language questions are later derived \cite{OmniSQL} (see right column in Figure \ref{overview}).

To systematically control and diversify the reasoning complexity of generated queries, we introduce a business-driven complexity control strategy.

Specifically, we define four business reasoning complexity levels. The design of these levels was reviewed by BI practitioners with experience in enterprise analytics, ensuring that each level corresponds to a realistic and commonly observed category of analytical questions in practice.

\begin{itemize}
    \item \textbf{Single Metric Analysis}: The simplest level, where questions request a single metric or a list of entities under straightforward filtering conditions (e.g., ``How many demo meetings did active SDRs book last week?'').

    \item \textbf{Comparative Metric Analysis}: Questions that compare the same metric across a single analytical dimension, such as time, region, or category (e.g., ``For the current quarter, what are the total outbound-sourced pipeline dollars created by region?'').

    \item \textbf{Derived Metric Analysis}: Questions that compute derived indicators or apply explicit business logic, such as ratios, percentages, or rule-based subsets (e.g., ``What percentage of primary partner co-sell opportunities are in Commit or already Closed-Won?'').

    \item \textbf{Compositional Task Analysis}: The highest complexity level, where questions combine multiple analytical operations within a single query, such as jointly comparing, ranking, and segmenting results (e.g., ``Which five region–product-family combinations generated the highest closed-won revenue, and what were their revenue totals?'').
\end{itemize}

By conditioning SQL generation to different levels of business reasoning complexity, we generate queries that cover various reasoning patterns. This enables systematic evaluation of Text-to-SQL agents under diverse challenging business tasks.

To ensure executability, we adopt an execution-based refinement loop to iteratively revise SQL queries based on database execution feedback, while preserving the original business logic and schema constraints \cite{OmniSQL,SQL-Factory}.

\paragraph{SQL-to-Question Conversion}\label{method_4}
After generating executable SQL queries, we convert each query into a semantically equivalent natural language question. 
The conversion is conditioned on the same structured persona-centric business logic used during SQL generation.

The question is phrased in business-oriented language, avoiding explicit references to tables, columns, or SQL constructs. In the meantime, the question captures the full analytical logic of the SQL query, including filters, aggregations, groupings, and constraints, using terminology appropriate to the persona.
This step produces aligned (question, SQL) pairs that are suitable for evaluating text-to-SQL systems in realistic BI scenarios.

\subsection{Data Quality Evaluation Metrics }\label{method_5}
To assess the quality of the synthesized Text-to-SQL data, we evaluate each (question, SQL) pair along two complementary dimensions that capture semantic correctness and business realism.

\textbf{Question--SQL Alignment} evaluates whether the SQL query correctly and completely answers the natural language question, focusing on semantic correctness. 
\textbf{Question Realism} evaluates whether the natural language question resembles realistic questions asked by professionals in real-world BI scenarios.
Each dimension is evaluated using a set of criteria summarized in Table~\ref{tab:eval_rubrics}. Full definitions of all evaluation criteria and guidelines are provided in Appendix~\ref{app:rubrics}.

Following prior work \cite{OmniSQL}, we implement all evaluations using an LLM-as-a-judge framework to enable scalable and consistent assessment. For each criterion, the judge assigns one of four ratings: \emph{excellent}, \emph{good}, \emph{average}, or \emph{poor}. Scores are aggregated into a single scalar using a weighted average:
\[
\text{Score} = \frac{1.0N_e + 0.75N_g + 0.5N_a + 0.25N_p}{N_e + N_g + N_a + N_p},
\]
where $N_e$, $N_g$, $N_a$, and $N_p$ denote the numbers of samples receiving each rating.

\begin{table}[t]
\centering
\begin{adjustbox}{width=\columnwidth}
\begin{tabular}{ll}
\toprule
\textbf{Evaluation Dimension} & \textbf{Criteria} \\
\midrule
Question--SQL Alignment & Result Completeness; Constraint Consistency; \\
& Structural Consistency; Unnecessary Complexity \\
Question Realism & Business Relevance; Business Language; \\
& Expression Naturalness; Decision Value \\
\bottomrule
\end{tabular}
\end{adjustbox}
\caption{Summary of data quality evaluation criteria.}

\label{tab:eval_rubrics}
\end{table}

\section{Experiments}
To demonstrate how our framework performs in the realistic enterprise BI setting, we conduct an experiment to generate  Text-to-SQL evaluation data in the sales analytics domain on a production-scale private Salesforce database.

\subsection{Database Settings}
The target database is a production-scale Salesforce database for sales analytics. It contains 710 tables, with the most complex table comprising 310 columns and 16 foreign-key relationships, resulting in a highly interconnected schema. The database follows Salesforce’s data model and includes core business entities such as leads, accounts, opportunities, activities, and their associated metadata.

Queries are executed using Salesforce Object Query Language (SOQL), a SQL-like query language with syntax and constraints that differ from standard relational SQL \cite{CRMArena}.

\subsection{Implementation Details}
Given the target sales domain, we first generate high-level functional departments (e.g., sales development, account management, pipeline management) to ensure coverage of the domain’s organizational structure. We define four seniority levels: individual contributor, manager, director, and executive, to capture variation in responsibility scope and decision-making authority. These seniority levels were reviewed by domain experts to ensure that they reflect common organizational structures in real-world business settings. Based on different departments and seniority levels, we generate \textbf{20 business personas} in the sales domain, each representing a realistic job role within the sales organization. Each persona is described by its job title, responsibility, primary goals, and pain points.

For each persona, we generate 5 persona-specific work scenarios representing responsibility areas (e.g., pipeline monitoring or performance review). Each scenario is further decomposed into a workflow of tasks and primary KPIs, yielding structured persona-centric business logic that mirrors how analytical questions arise in the sales domain.

The generated personas, work scenarios, and workflows were reviewed by a sales domain BI expert to validate that the modeled business logic reflects realistic sales roles and workflows.  

\begin{figure}[t]
  \includegraphics[width=\columnwidth]{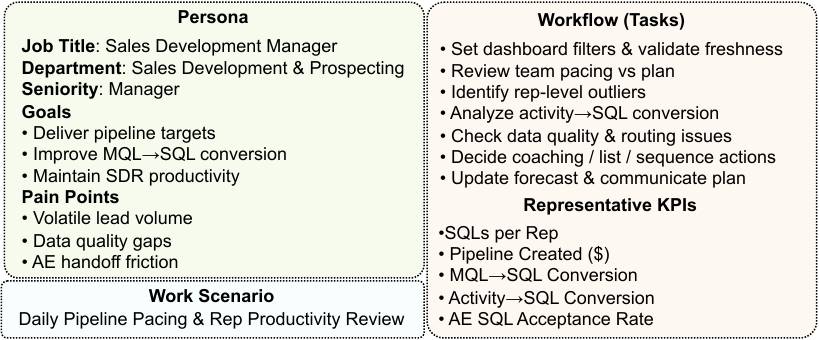}
  \caption{An example structured persona-centric business logic in the sales domain.}
  \label{fig:persona}
\end{figure}

To generate Text-to-SQL evaluation data, we select 20 validated business logic instances, each consisting of one persona paired with one work scenario and its associated workflow. 
(An example instance structure is shown in Figure \ref{fig:persona}). These business logic instances are used to guide schema selection and data generation.
We employ GPT-5\cite{gpt} for both business logic modeling and data synthesis. All prompts are listed in Appendix~\ref{app:prompts}. In total, we generate \textbf{240 question–SQL pairs}, spanning diverse personas and business reasoning complexity levels.

\input{latex/data_quality_eval_tab}

\section{Results}
To evaluate the quality of the generated dataset, we randomly sample 100 instances spanning all business reasoning complexity levels from our dataset. For comparison, we apply OmniSQL~\cite{OmniSQL} and SQL-Factory~\cite{SQL-Factory} using their publicly released implementations, and adapt them to support the Salesforce database, SOQL dialect, and other experimental settings. 
To mitigate evaluator bias, we employ two independent judging models: Gemini-3-Pro~\cite{gemini} and Claude Sonnet~4.5~\cite{claude}. The evaluation results are summarized in Table~\ref{tab:data_quality_eval}.

\paragraph{Comparison with SOTA Methods}
Using both judging LLMs, our method substantially outperforms prior methods on question realism,  while maintaining strong question--SQL alignment.
Under Gemini-3-Pro evaluation, our method achieves a question realism score of 98.44\%, outperforming OmniSQL by +19.5\% and SQL-Factory by +54.7\%. Similar gains are observed with Claude Sonnet~4.5, where our method exceeds OmniSQL by +16.4\% and SQL-Factory by over +50\% in question realism.
At the same time, our method maintains consistently high question--SQL alignment, achieving 98.59\% under Gemini-3-Pro and 90.94\% under Claude Sonnet~4.5, which are comparable to or higher than OmniSQL and SQL-Factory.
These results indicate that grounding data generation in business logic significantly improves the business realism of generated questions, whereas existing methods that prioritize schema coverage and SQL pattern diversity tend to produce generic queries weakly connected to real-world business logic.

\input{latex/data_quality_eval_inner_method_tab}
\paragraph{Data Quality Analysis}
We observe that Claude Sonnet~4.5 is stricter than Gemini-3-Pro across both evaluation metrics. Therefore, we select Claude Sonnet~4.5 as the evaluator for further analysis. We evaluate all 240 samples generated by our framework and compare data quality across different business reasoning complexity levels (Table~\ref{tab:data_quality_eval_inner_method}).

In addition to question--SQL alignment and question realism, we add another metric, \emph{question--persona alignment}, to evaluate whether a generated question is consistent with the persona used to guide data generation. Detailed evaluation criteria are provided in Appendix~\ref{app:rubrics}.

As the complexity level increases from single metric analysis to compositional task analysis, the question--SQL alignment decreases from 95.61\% to 90.22\%, reflecting the increasing analytical difficulty of the queries. Question realism remains comparatively robust, declining only modestly from 97.80\% to 95.11\%, which indicates that the generated questions remain business realistic even for complex analytical tasks. Persona--question alignment exhibits a larger drop, from 95.27\% to 85.46\%, suggesting that maintaining explicit persona consistency becomes more challenging as queries encode more abstract and compositional business logic. Nevertheless, persona alignment remains strong overall, and the limited degradation in question realism indicates that the dataset preserves business realism across all complexity levels.

\paragraph{Human Evaluation}
To further validate data quality, we conduct a human evaluation using the 100 data samples. The evaluation is performed by three senior graduate students with backgrounds in computer science and active research experience in Text-to-SQL systems.
Following prior work \cite{OmniSQL}, to reduce annotation burden while maintaining reliability, annotators provide binary pass/fail judgments along two dimensions: Question–SQL Alignment and Question Realism.

Each sample is evaluated independently, with disagreements resolved through discussion. The results show that 89\% of the evaluated samples pass the question–SQL alignment criterion, and over 98\% are judged to be realistic. These findings indicate that the majority of the synthesized data achieves high-quality in both semantic correctness and business realism.

\section{Error Analysis}
We observe that question–persona alignment consistently declines as business reasoning complexity level increases, despite questions remaining business-realistic (see in Table \ref{tab:data_quality_eval_inner_method}).
This degradation arises because higher-complexity questions increasingly combine different analytical tasks that are owned by different personas. Questions typically align with a persona’s core scope when the question complexity is low (e.g., a compensation analyst validating payout totals or an account executive checking engagement signals). However, as the complexity increases, questions tend to combine multiple analytical tasks that are normally handled by different personas. 

For example, a question generated for a Sales Compensation Analyst asks: \textit{For this month’s commission run, which reps have the highest Closed Won ACV from accounts with contracts expiring in the next 60 days?} While commission runs and Closed Won ACV fall within the analyst’s scope, contract expiration analysis and account scoping are typically owned by Account Management or Sales Operations. 
As a result, increasing reasoning complexity leads models to combine analytical tasks without respecting persona boundaries, producing questions that are business-plausible but misaligned with persona-specific workflows.
However, such cross-functional questions can naturally arise in real-world BI settings, particularly for senior roles or exploratory analysts, and highlight the challenge of maintaining strict persona alignment under complex analytical tasks.

\input{latex/ablation_study_tab}

\section{Ablation Study}
We conduct an ablation study to quantify the contribution of two core components of our framework: (i) structured business logic guidance and (ii) business reasoning complexity control. Results are summarized in Table~\ref{tab:ablation}.

\paragraph{Ablation Settings}
To isolate the effect of each component, we consider two ablated variants. In \textit{w/o BL}, we remove the Business Logic Modeling module and randomly select 1--3 tables as input for SQL generation, while retaining the business reasoning complexity control strategy. In \textit{w/o BL, w/o CC}, both business logic and complexity control are removed, and SQL generation is conditioned solely on randomly selected table schema.

\paragraph{Complexity Diversity Metric}
We measure the effectiveness of business reasoning complexity control using a \emph{complexity diversity} metric, which quantifies how generated samples are distributed across the four predefined business reasoning complexity levels (Section~\ref{method_3}). Each (question, SQL) pair is assigned a complexity label by Claude Sonnet~4.5. Complexity diversity is computed using normalized Shannon entropy~\cite{entropy}:
\[
H_{\text{complexity\_diversity}} = \frac{- \sum_{k=1}^{K} p_k \log p_k}{\log K},
\]
where \( p_k \) denotes the proportion of samples at complexity level \( k \), and \(K\) is the total number of predefined complexity levels (with \(K=4\) in our setting, see Section \ref{method_3}). Higher values indicate more balanced coverage of reasoning complexity levels.

\paragraph{Impact of Business Logic Guidance}
Removing business logic (\textit{w/o BL}) guidance causes a substantial degradation in question realism, which drops from 97.34\% to 63.38\%. This result indicates that schema-driven generation without explicit business context leads to questions that are detached from realistic business logic. 

\paragraph{Impact of Complexity Control}
Removing Business Reasoning Complexity Control (\textit{w/o BL, w/o CC}) causes complexity diversity to drop sharply from 0.91 to 0.54, indicating that schema-only generation lacks systematic coverage of diverse business reasoning complexity. 
Notably, removing business logic (\textit{w/o BL}) achieves a marginally higher complexity diversity score (0.91) than the full framework (0.85). We hypothesize that this difference arises from interactions between business logic constraints and complexity control levels.

\paragraph{Alignment vs. Realism}
Across all settings, question--SQL alignment remains consistently high (above 90\%), suggesting that semantic correctness alone is insufficient to assess data quality in BI settings. The sharp divergence between alignment and realism scores further underscores the necessity of explicit business logic modeling to produce Text-to-SQL data that is both semantically correct and business realistic.

\input{latex/section/sota_model_performance}

\section{Conclusion}
We proposed a \textbf{Business Logic–Driven Data Synthesis} framework to generate business-realistic Text-to-SQL evaluation data in private BI settings. By explicitly grounding data generation in business personas, work scenarios, and workflows, the synthesized data reflects real-world business logic. In addition, we introduce a \textbf{business reasoning complexity control} strategy to control and diversify the business reasoning steps and patterns of the synthesized data samples.
Experiments on a production-scale enterprise database demonstrate that the synthesized data achieves high business realism and strong semantic correctness.

In future work, we plan to extend our framework to support multi-turn, interactive analytical tasks that more closely reflect how business users iteratively explore data in real BI workflows.

\section*{Limitations}

Our framework is designed for business intelligence (BI) scenarios and does not aim to address all Text-to-SQL tasks. It relies on structured business logic, such as personas, work scenarios, and workflows, which naturally align with the BI domain but may be less applicable to other settings.

In addition, we validate our framework primarily in the sales domain using a production-scale Salesforce database. While the sales domain represents a complex and representative BI use case, further validation is needed to assess the generalizability of our approach across other BI domains, such as finance, marketing, or operations.

\section*{Potential Risks}

Although our framework operates on a private enterprise schema and generates synthetic questions without exposing raw data values, there remains a risk that schema-level information, business terminology, or structural patterns could reveal sensitive organizational practices if datasets or prompts are improperly shared.

\bibliography{custom}

\clearpage
\appendix
\input{latex/section/appendix}

\end{document}

%% file: latex/section/related_work.tex
\paragraph{Text-to-SQL Benchmarks}
Text-to-SQL research has been driven by public benchmarks that emphasize cross-domain generalization and compositional SQL reasoning. Spider~\cite{yu-etal-2018-spider} established a large cross-domain setting with complex multi-table queries. BIRD~\cite{BIRD} introduces large-scale databases with noisy values, external knowledge requirements, and efficiency constraints. Some benchmarks further expand realism: Spider~2.0~\cite{spider2} targets enterprise-scale workflows and multi-dialect SQL, BIRD-Interact~\cite{BIRD-INTERACT} evaluates multi-turn interactive Text-to-SQL behavior, and LiveSQLBench~\cite{livesqlbench2025} provides a continuously updated, contamination-resistant testbed reflecting real-world SQL usage. While these open benchmarks provide strong and increasingly realistic testbeds, evaluating Text-to-SQL agents in private BI enterprise databases remains challenging due to domain specificity, proprietary schema, and business-specific workflows.

\paragraph{Business-Oriented Benchmarks}
Recent works have increasingly focused on evaluating agents in realistic professional tasks. $\tau$-bench~\cite{taubench} benchmarks tool-augmented agents in real-world domains, emphasizing robustness and consistency in task completion. WorkArena~\cite{WorkArena} and its extension WorkArena++~\cite{WorkArena++} evaluate agents on common enterprise knowledge-work tasks within realistic software environments, highlighting challenges in planning, reasoning, and workflow execution.
More closely related to business analytics, CRMArena~\cite{CRMArena} introduces a Salesforce-like database sandbox and evaluates agents on expert-validated professional analytical tasks, while CRMArena-Pro~\cite{CRMArena-Pro} expands coverage across diverse business scenarios and interaction patterns. These efforts motivate benchmarks and data generation frameworks that ground Text-to-SQL agent evaluation in realistic business contexts.

\paragraph{Synthetic Data Generation for Text-to-SQL}
To mitigate the scarcity and cost of human-labeled Text-to-SQL datasets, a growing body of work has explored synthetic data generation. OmniSQL~\cite{OmniSQL} introduces a scalable synthesis framework emphasizing broad schema and query pattern coverage. SQL-Factory~\cite{SQL-Factory} formulates synthesis as a multi-agent system that improves SQL diversity and quality through role specialization and iterative refinement. Beyond direct SQL generation, several approaches incorporate intermediate representations to improve controllability and realism. For example, SQLBarber~\cite{SQLBarber} generates customized SQL workloads under user-specified constraints for realistic benchmarking, while DSQG-Syn~\cite{dsqg-syn} incorporates domain-specific question types to improve coverage within a target domain. 

Despite these advances, existing synthesis methods primarily optimize for SQL pattern diversity and schema coverage while neglecting business realism. As a result, the generated questions are often detached from realistic business logic and therefore fail to support reliable evaluation of Text-to-SQL agents in BI settings.

%% file: latex/data_quality_eval_tab.tex
\begin{table}[t]
\centering
\begin{adjustbox}{width=\columnwidth}
\begin{tabular}{c c l l}
\toprule
\textbf{Method} & \textbf{Evaluation LLM} 
& \makecell{\textbf{Question-SQL} \\ \textbf{Alignment}(\%)}
& \makecell{\textbf{Question} \\ \textbf{Realism}(\%)} \\
\midrule
\multirow{2}{*}{\makecell{OmniSQL \\ \cite{OmniSQL}}}
& Gemini3 Pro 
& 95.47 \,$(\downarrow 3.12)$
& 78.91 \,$(\downarrow 19.53)$ \\
& Claude Sonnet 4.5 
& 88.59 \,$(\downarrow 2.35) $
& 80.94 \,$(\downarrow 16.40)$ \\
\midrule
\multirow{2}{*}{\makecell{SQL-Factory \\ \cite{SQL-Factory}}} 
& Gemini3 Pro 
& 90.40 \,$(\downarrow 8.19) $
& 43.75 \,$(\downarrow 54.69)$ \\
& Claude Sonnet 4.5 
& 84.15 \,$(\downarrow 6.79) $
& 45.43 \,$(\downarrow 51.91)$\\
\midrule
\multirow{2}{*}{Ours} 
& Gemini3 Pro 
& \textbf{98.59} 
& \textbf{98.44} \\
& Claude Sonnet 4.5 
& \textbf{90.94} 
& \textbf{97.34} \\
\bottomrule
\end{tabular}
\end{adjustbox}
\caption{Data quality comparison with existing Text-to-SQL data generation methods.}
\label{tab:data_quality_eval}
\end{table}

%% file: latex/data_quality_eval_inner_method_tab.tex
\begin{table}
\centering
\begin{adjustbox}{width=\columnwidth}
\begin{tabular}{lccc}
\toprule
\textbf{Complexity Level} 
& \makecell{\textbf{Question-SQL} \\ \textbf{Alignment}(\%)}
& \makecell{\textbf{Question} \\ \textbf{Realism}(\%)}
& \makecell{\textbf{Question-Persona} \\ \textbf{Alignment}(\%)} \\
\midrule
Single Metric.      & \textbf{95.61} & \textbf{97.80} & \textbf{95.27} \\
Comparative Metric.  & 88.75 & 96.79 & 94.91 \\
Derived Metric.      & 90.38 & 96.88 & 88.97 \\
Compositional task.       & 90.22 & 95.11 & 85.46 \\
\bottomrule
\end{tabular}
\end{adjustbox}
\caption{Fine-grained data quality evaluation across business reasoning complexity levels, reporting overall scores for each evaluation dimension.}
\label{tab:data_quality_eval_inner_method}
\end{table}

%% file: latex/ablation_study_tab.tex
\begin{table}
\centering
\begin{adjustbox}{width=\columnwidth}
\begin{tabular}{l|ccc}
\toprule

& \makecell{\textbf{Question-SQL} \\ \textbf{Alignment}(\%)}
& \makecell{\textbf{Question} \\ \textbf{Realism}(\%)}
& \makecell{\textbf{Complexity} \\ \textbf{Diversity}(\%)} \\
\midrule
ours  
& 90.94
& \textbf{97.34}
& 0.85
\\
w/o BL
& 94.63
& 63.38
& \textbf{0.91}
\\
w/o BL, w/o CC
& \textbf{94.67}
& 48.55
& 0.54
\\
\bottomrule
\end{tabular}
\end{adjustbox}
\caption{Ablation study on Business Logic Guidance (BL) and Business Reasoning Complexity Control (CC).}
\label{tab:ablation}
\end{table}

%% file: latex/section/sota_model_performance.tex
\section{SOTA Model Performance on Our Synthetic Data}

To demonstrate that the synthesized dataset is both high-quality and challenging, we evaluate the state-of-the-art Text-to-SQL agents on the dataset and analyze the agent performance across different business reasoning complexity levels.

\subsection{Experimental Setup}
We evaluate a set of representative proprietary and open-source LLMs, together with a ReAct-style agent following the CRMArena implementation \cite{CRMArena}. All models are evaluated on a subset of 110 high-quality samples verified by our LLM-as-a-judge data quality evaluator (Section~\ref{method_5}). The selected samples achieve a perfect score in both Question-SQL Alignment and Question Realism metrics with a balanced distribution across the four business reasoning complexity levels. Performance is measured using execution accuracy (EX), following prior works \cite{yu-etal-2018-spider,BIRD}. 
Full experimental details and model lists are provided in Appendix~\ref{app:sota-agent}. 

\input{latex/agent_eval_tab}
\subsection{Results and Analysis}
Table~\ref{main_results} summarizes execution accuracy across complexity levels. Overall, even the strongest models exhibit substantial performance degradation on most complex business-driven queries. Excluding GPT-5 (which is also used during data generation), the best-performing model achieves only 65.79\% overall accuracy and 42.86\% on the compositional task analysis level. This result indicates that the synthesized dataset is sufficiently challenging and not trivially solvable by current models.

More importantly, performance consistently declines as business reasoning complexity increases across all evaluated models. This trend aligns with our business-driven complexity design, that the dataset effectively differentiates models based on their ability to handle multi-step, complex business reasoning rather than SQL patterns.

%% file: latex/agent_eval_tab.tex
\begin{table}[t]
  \centering
  \begin{adjustbox}{width=\columnwidth}
    \begin{tabular}{l|cccc|c}
      \toprule
      \textbf{Model} 
        & \makecell{\textbf{Single} \\ \textbf{Metric.}}
        & \makecell{\textbf{Comparative} \\ \textbf{Metric.}}
        & \makecell{\textbf{Derived} \\ \textbf{Metric.}}
        & \makecell{\textbf{Compositional} \\ \textbf{Task.}}
        & \textbf{Overall} \\
      \midrule

      \multicolumn{6}{c}{\textit{Proprietary Models}} \\
      \midrule
      GPT-5                  & \textbf{90.24} & \textbf{70.97} & \textbf{80.95} & \textbf{76.19} & \textbf{80.70} \\
      Claude-4.5-Sonnet      & 73.17 & 48.39 & 57.14 & 28.57 & 55.26 \\
      Gemini-3-Pro           & 65.85 & \textbf{67.74} & \textbf{76.19} & 33.33 & 62.28 \\
      Gemini-2.5-Pro         & \textbf{75.61} & 64.52 & 71.43 & \textbf{42.86} & \textbf{65.79} \\
      \midrule

      \multicolumn{6}{c}{\textit{Open-Source Models}} \\
      \midrule
      DeepSeek-v3.2          & 58.54 & 25.81 & 38.10 & 28.57 & 40.35  \\
      DeepSeek-R1            & 63.41 & \textbf{41.94} & 42.86 & \textbf{33.33} & \textbf{48.25}  \\
      Qwen3-235B             & \textbf{70.73} & 29.03 & \textbf{47.62} & 28.57 & 47.37  \\
      Qwen3-30B              & 68.29 & 29.03 & 38.10 & 19.05 & 42.98   \\
      Qwen3-Coder-30b        & 51.22 & 25.81 & 42.86 & 23.81 & 37.72  \\
      \bottomrule
    \end{tabular}
  \end{adjustbox}
  \caption{\label{main_results}
    Results of state-of-the-art Text-to-SQL models on the synthesized dataset. GPT-5 is also used during data generation, which may partially explain its comparatively higher performance.
  }
  \label{main_results}
\end{table}

%% file: latex/section/appendix.tex

\section{Detailed Evaluation Rubrics}
\label{app:rubrics}
To assess the quality of the synthesized dataset, we conduct systematic data quality evaluation along three complementary dimensions: \emph{question--SQL alignment}, \emph{question realism}, and \emph{persona--question alignment}. These dimensions capture semantic correctness, business plausibility, and persona grounding of the generated Text-to-SQL pairs.

\paragraph{Question-SQL Alignment.}
Question-SQL alignment measures whether a natural language question is semantically and operationally answered by its paired SQL query. This dimension evaluates correctness from the perspective of downstream Text-to-SQL usage, where mismatches between intent and execution can lead to incorrect or misleading results. We decompose alignment into four criteria.
(1) \textbf{Result Completeness} assesses whether the SQL query returns exactly the information requested by the question, without omitting required outputs or introducing irrelevant ones.  
(2) \textbf{Constraint Consistency} evaluates whether filters, conditions, and qualifiers expressed or implied in the question are faithfully reflected in the SQL logic.  
(3) \textbf{Structural Consistency} examines whether the query structure---including aggregations, groupings, joins, and calculations---matches the analytical intent of the question.  
(4) \textbf{Unnecessary Complexity} penalizes unnecessary joins, filters, subqueries, or computations that are not required by the question and may obscure or alter its intended meaning.


\paragraph{Question Realism.}
Question realism evaluates whether generated questions resemble those that business users would plausibly ask in real-world BI settings. Rather than focusing on syntactic correctness, this dimension measures naturalness, motivation, and practical usefulness. We assess realism using four criteria.
(1) \textbf{Business Relevance} measures whether a question reflects a genuine business concern, decision, or workflow, as opposed to an artificial or purely exploratory query.  
(2) \textbf{Business Language} evaluates whether the question employs domain-appropriate terminology, metrics, and concepts in a manner consistent with professional practice.  
(3) \textbf{Expression Naturalness} assesses whether the question is expressed in clear, fluent, and human-like language, rather than rigid or template-driven phrasing.  
(4) \textbf{Decision Value} measures whether answering the question would support concrete actions, prioritization, or follow-up analysis in a business context.

\paragraph{Persona-Question Alignment.}
Because our synthesis framework explicitly conditions query generation on business personas and work scenarios, we additionally evaluate whether the generated questions are aligned with the intended persona context. Persona-question alignment assesses whether a question is appropriate for a given job role and naturally arises within a specific work scenario. We evaluate alignment along four aspects: 
(1) \textbf{role responsibility}, which checks whether the question falls within the typical decision scope of the persona; 
(2) \textbf{scenario relevance}, which measures whether the question plausibly arises while carrying out the given work scenario; 
(3) \textbf{practical value}, which evaluates whether answering the question would support decisions, actions, or monitoring in that context; and 
(4)\textbf{role language}, which assesses whether the phrasing reflects how someone in this role would typically think and communicate.

\section{SOTA Model Performance Experiment Details}\label{app:sota-agent}
\subsection{Experimental Settings}
\paragraph{Models} We evaluate a diverse set of proprietary and open-source LLMs, including: GPT family (GPT-5)\cite{gpt}, Claude family (Claude-4.5-Sonnet)\cite{claude}, Gemini family (Gemini-3-Pro, Gemini-2.5-Pro)\cite{gemini}, DeepSeek family (deepseek-v3.2, deepseek-r1), and Qwen family (Qwen3-235B\footnote{Qwen/Qwen3-235B-A22B-Instruct-2507},
 ,Qwen3-Coder-30b\footnote{Qwen/Qwen3-Coder-30B-A3B-Instruct}, and Qwen3-30B\footnote{Qwen/Qwen3-30B-A3B-Instruct-2507}\cite{qwen3}.
These models represent a range of architectures, training scales, and reasoning capabilities, enabling a broad comparison across contemporary Text-to-SQL systems.

\paragraph{ReAct-Based Agent}
We adopt a prompt-based agentic framework following the ReAct paradigm \cite{ReAct}, using the implementation provided in CRMArena \cite{CRMArena}. Each interaction step consists of a thought and an action. The action can be one of two types: (1) \textbf{execute}, which generates and executes a SOQL query against the database and returns the execution result as an observation, and (2) \textbf{respond}, which outputs the agent’s final answer for the task. To simplify the experimental setting, we omit an explicit schema-linking step and instead provide the agent with a fixed schema consisting of all tables appearing in the selected dataset. This schema includes 30 tables and is directly inserted into the prompt. Execution results are returned as observations and used to guide subsequent reasoning steps. The agent terminates when it issues a response action or reaches a predefined maximum steps.

\paragraph{Dataset and Metrics}
All models are evaluated on the synthesized dataset generated by our framework. To ensure high data quality, we apply the LLM-as-a-judge procedure described in Section~\ref{method_5} as a post-generation verification step. Only samples receiving perfect scores across all evaluation criteria are retained for benchmarking. From the verified set, we randomly select 110 samples, following a balanced ratio across the four complexity levels. Performance is measured using the widely adopted metric of \textbf{execution accuracy (EX)} \cite{yu-etal-2018-spider,BIRD}, defined as the percentage of queries whose execution results match the expected outputs.

\section{Prompts}\label{app:prompts}

\subsection{Persona Modeling Prompt}
{\small
\begin{verbatim}
You are a senior business intelligence consultant and
organizational workflow analyst.

Your task is to generate realistic job personas for a
specific functional area within a business domain.

A persona represents a professional role with concrete
responsibilities, goals, challenges, and success metrics.
Personas should reflect how real people work in practice.

Personas must:
- Fit naturally within the given functional area
- Span all seniority levels (not manager-heavy)
- Reflect real workflows, including communication,
  decision-making, planning, execution, coordination,
  and use of data
- Be distinct from each other
- Represent roles commonly found in organizations
  operating in this domain

You MUST output valid JSON only.
Do NOT include any text outside the JSON object.

Generate {{num_personas}} distinct personas.

DOMAIN: {{domain_name}}

FUNCTIONAL AREA:
- Name: {{functional_area_name}}
- Description: {{functional_area_desc}}

Coverage requirements:
- Personas MUST cover all seniority levels
- Personas must reflect different responsibilities or
  perspectives within the functional area

For each persona, output:
- name: realistic job title
- seniority_level:
  individual_contributor | manager |
  director | executive
- role_summary:
  1–3 sentences describing actual work performed
- primary_goals:
  role objectives or desired outcomes
- pain_points:
  recurring challenges or constraints
- primary_kpis:
  metrics used to evaluate success

Output JSON with the following structure:

{
  "personas": [
    {
      "name": "string",
      "seniority": "individual_contributor 
      | manager | director | executive",
      "role_summary": "string",
      "primary_goals": ["string"],
      "pain_points": ["string"],
      "primary_kpis": ["string"]
    }
  ]
}
\end{verbatim}
}

\subsection{Work Scenario Modeling Prompt}
{\small
\begin{verbatim}
You are a senior organizational workflow analyst
specializing in modeling real-world work practices.

You are given a persona (job role) within a specific
functional area of a business domain.

Your task is to identify the major work scenarios the
persona encounters in day-to-day responsibilities.

A work scenario is a recurring situation or context
in which the persona performs actions, makes decisions,
communicates with others, analyzes information,
handles exceptions, or completes processes.

Work scenarios must be:
- Realistic for the domain
- Comprehensive enough to reflect the full scope
  of the persona’s responsibilities
- Balanced across communication, coordination,
  analysis, planning, execution, and other
  domain-relevant behaviors

You MUST output valid JSON only.
Do NOT include any text outside the JSON object.

DOMAIN:
{{domain_name}}

FUNCTIONAL AREA:
- Name: {{functional_area_name}}
- Description: {{functional_area_desc}}

PERSONA:
{{persona}}

For this persona, generate work scenarios that:
- Represent key recurring responsibility areas
- Include both routine and non-routine aspects
- May involve people, tools, systems, or data
- Require decision-making, planning, or analysis
- Vary in timescale (e.g., daily, weekly,
  monthly, quarterly, ad-hoc)
- Are appropriate for the persona’s seniority
  and role responsibilities

Generate 3–5 distinct work scenarios in which
the persona uses data or analytics.

Each scenario must include:
- name: short descriptive title
- description: 1–3 sentences
- frequency: typical occurrence
  (e.g., daily, weekly, monthly, quarterly, ad-hoc)

Output JSON with the following structure:

{
  "work_scenarios": [
    {
      "persona_name": "string",
      "work_scenarios": [
        {
          "name": "string",
          "description": "string",
          "frequency": "string"
        }
      ]
    }
  ]
}

\end{verbatim}
}

\subsection{Workflow Modeling Prompt}
{\small
\begin{verbatim}
You are a senior business workflow architect.

You are given:
- A functional area within a business domain
- A specific persona (job role)
- One work scenario for that persona

Your task is to describe the detailed tasks the
persona performs in this work scenario.

A task is a meaningful step in the workflow.
Tasks should:
- Form a coherent sequence (not independent bullets)
- Reflect how a real person completes the work
  from start to finish
- Include actions, communication, coordination,
  information checking, or decision-making
- Represent practical day-to-day behavior,
  not idealized processes

You MUST output valid JSON only.

DOMAIN:
{{domain_name}}

FUNCTIONAL AREA:
- Name: {{functional_area_name}}
- Description: {{functional_area_desc}}

PERSONA:
{{persona}}

WORK SCENARIO:
{{scenario}}

TASK GENERATION REQUIREMENTS:

For this (persona, scenario), generate:

1) Tasks
- Generate 5–8 ordered tasks forming the workflow
- For each task:
  - name: short action phrase
  - description: 1–3 sentences explaining
    what is done and why

2) Related KPIs
- Generate 5–15 KPIs relevant to this scenario
- For each KPI:
  - name
  - description (1–3 sentences)
  - formula_or_logic: high-level computation logic

Output JSON with the following structure:

{
  "tasks": [
    {
      "name": "string",
      "description": "string"
    }
  ],
  "related_kpis": [
    {
      "name": "string",
      "description": "string",
      "formula_or_logic": "string"
    }
  ]
}

\end{verbatim}
}

\subsection{Business Logic-Driven Schema Selection Prompt}
{\small
\begin{verbatim}
You are an expert data engineer and business
intelligence analyst.

You are helping a given persona decide which
database tables are relevant for one of their
recurring work scenarios.

Your task is to rate the relevance of each
database table for this specific scenario,
given the persona, their workflow tasks,
the business entities involved, and the KPIs
they care about.

Scoring rules (per table):
- 3 = Essential
      Clearly required for core analyses or
      operations in this scenario
- 2 = Useful / relevant
      Often joined or used as supporting
      context or attributes
- 1 = Possibly related
      Marginally relevant in limited cases
- 0 = Irrelevant
      No meaningful connection to this scenario

You MUST output valid JSON only.

DOMAIN:
{{DOMAIN_NAME}}

FUNCTIONAL AREA:
- Name: {{functional_area_name}}
- Description: {{functional_area_desc}}

PERSONA:
{{persona_block}}

WORK SCENARIO:
{{scenario_block}}

WORKFLOW TASKS:
Below are the main steps the persona performs
in this scenario:
{{tasks_block}}

DATABASE TABLES TO SCORE:
Below is a list of database tables with brief
schema or column summaries.

For EACH table:
- Assign a relevance score (0–3)
- Provide a short justification referring to
  the scenario, tasks, entities, or KPIs

{{tables_block}}

Guidance:
- Use score 3 for tables that directly store
  core entities, events, or measurements used
  in the tasks or KPIs (e.g., fact tables)
- Use score 2 for tables commonly joined to
  add attributes or context (e.g., dimensions)
- Use score 1 for weak or occasional relevance
- Use score 0 for no meaningful connection

Return JSON in the following format ONLY:

{
  "scores": [
    {
      "table": "table_name",
      "score": 0,
      "reason": "short explanation"
    }
  ]
}

\end{verbatim}
}

\subsection{Business-Driven SQL Generation Prompt}
{\small
\begin{verbatim}
You are an expert data analyst and {{db_engine_name}}
query engineer.

You are given:
- A target business intent complexity level
- A business domain and functional area
- One persona and one of their work scenarios
- Workflow tasks for this scenario
- A subset of database tables and columns

Your task is to generate realistic analytic SQL
queries that this persona would use in this
scenario, matching the specified INTENT
COMPLEXITY.

INTENT COMPLEXITY
{{complexity_block}}

Each query MUST reflect the required business
reasoning depth. Complexity is defined by the
implicit reasoning steps needed to answer the
intent, NOT by SQL syntax alone.

CONTEXT
DOMAIN:
{{DOMAIN_NAME}}

FUNCTIONAL AREA:
- Name: {{functional_area_name}}
- Description: {{functional_area_desc}}

PERSONA:
{{persona_block}}

WORK SCENARIO:
{{scenario_block}}

WORKFLOW TASKS:
{{tasks_block}}

KEY KPIS:
{{kpis_block}}

ALLOWED SCHEMA (TABLES AND COLUMNS):
You may ONLY use the following tables/columns:
{{tables_block}}

TASK
Generate {{num_queries}} diverse analytic queries.

For each query:
- Choose a realistic business intent aligned with:
  - one or more workflow tasks, OR
  - a KPI, OR
  - a decision the persona must make
- Ensure the intent requires the specified
  complexity level
- Write a valid {{db_engine_name}} SQL query
  using ONLY the allowed schema

Rules:
- Use ONLY the provided tables and columns
- Do NOT invent tables, columns, or values
- Queries MUST match the INTENT COMPLEXITY
- Queries must be realistic BI queries
- Each query must differ meaningfully in intent,
  focus, or analysis
- Output MUST be valid JSON only

Output JSON format ONLY:

{
  "queries": [
    {
      "intent_complexity": 
      "{{intent_complexity_level}}",
      "intent": "string",
      "sql": "string"
    }
  ]
}

\end{verbatim}
}

\subsection{Execution-based SQL Refinement Prompt}
{\small
\begin{verbatim}
You are an expert data analyst and {{db_engine_name}}
query engineer.

You are given:
- A business domain
- A persona and one work scenario
- Workflow tasks for this scenario
- An allowed schema subset (tables + columns)
- A prior SQL attempt for {{db_engine_name}}
- An execution error message

Your task is to FIX the SQL query.

Rules:
- Do NOT invent tables, columns, or values
- Preserve the business intent as much as possible
- Output MUST use valid {{db_engine_name}} syntax
- Verify referenced tables/columns exist in the
  provided schema
- Keep the corrected query close to the original
  structure unless the error requires a rewrite
- Output MUST be valid JSON only

DOMAIN:
{{DOMAIN_NAME}}

FUNCTIONAL AREA:
- Name: {{functional_area_name}}
- Description: {{functional_area_desc}}

PERSONA:
{{persona_block}}

WORK SCENARIO:
{{scenario_block}}

WORKFLOW TASKS:
Main steps the persona performs in this scenario:
{{tasks_block}}

SCENARIO KPIs:
{{kpis_block}}

ALLOWED SCHEMA (TABLES AND COLUMNS):
You may ONLY use the following tables/columns:
{{tables_block}}

INTENT:
{{intent_block}}

PREVIOUS SQL:
{{sql_block}}

ERROR MESSAGE:
{{error_block}}

TASK:
Rewrite the SQL query so that:
- It answers the same intent
- It uses only allowed tables/columns
- It is valid {{db_engine_name}} syntax
- It fixes the reported error

Output JSON format ONLY:

{
  "reasoning": "short explanation of changes",
  "sql": "corrected SQL query"
}

\end{verbatim}
}

\subsection{SQL-to-Question Conversion  Prompt}
{\small
\begin{verbatim}
You are an expert business data analyst and
{{db_engine_name}} query engineer.

You will be given:
- A business domain
- A specific persona (job role)
- One work scenario for that persona
- Detailed workflow tasks
- Key business entities and KPIs
- A subset of database tables and columns
- One SQL query using only the allowed schema

Your task is to write ONE natural-language
QUESTION that a real business user—specifically,
this persona in this scenario—might ask, and
that is answered exactly by the SQL query.

DOMAIN:
{{DOMAIN_NAME}}

FUNCTIONAL AREA:
- Name: {{functional_area_name}}
- Description: {{functional_area_desc}}

PERSONA:
{{persona_block}}

WORK SCENARIO:
{{scenario_block}}

WORKFLOW TASKS:
{{tasks_block}}

KEY KPIS:
{{kpis_block}}

ALLOWED SCHEMA (TABLES AND COLUMNS):
{{tables_block}}

SQL QUERY INTENT (HINT):
{{intent}}

SQL QUERY:
{{sql_text}}

TASK:
Write ONE natural-language QUESTION that:

1) Is answered exactly by the SQL query
2) Mentions ONLY concepts, filters, segments,
   time ranges, and conditions that are
   explicitly present in the SQL logic
3) Reflects the output shape of the SQL:
   - Grouped results → ask for a breakdown
   - Ranked or limited results → reflect ranking
   - Single aggregate → ask for that value
4) Reflects:
   - The persona’s goals and responsibilities
   - The scenario context
   - The workflow tasks
   - The entities and KPIs
   - The actual logic of the SQL query

Requirements:
- Use domain and business terminology implied
  by the persona, scenario, tasks, entities,
  and KPIs
- Do NOT mention SQL, tables, columns, joins,
  or technical details
- The question must sound realistic for
  this persona
- Use the SQL intent only as a hint; improve
  clarity and realism as needed
- Output MUST be valid JSON only

Return JSON ONLY:

{
  "question": "business natural language question",
  "paraphrases": [
    "alternate phrasing 1",
    "alternate phrasing 2"
  ]
}

\end{verbatim}
}

\subsection{LLM-as-a-Judge Prompt--Question-SQL Alignment}
{\small
\begin{verbatim}
ROLE
You are a data science expert and
{{db_engine_name}} query engineer.

Your task is to perform a strict semantic
audit between a natural-language question
and a generated SQL query.

CONTEXT
BUSINESS DOMAIN:
{{DOMAIN_NAME}}

DATABASE SCHEMA:
{{schema_str}}

INPUT
QUESTION:
{{question}}

SQL QUERY:
{{sql}}

EVALUATION GOAL
Determine whether the SQL logic accurately
translates the business intent of the
natural-language question, given the
provided schema.

EVALUATION CRITERIA

1) Result Completeness
Does the SQL return exactly the information
requested in the question, without missing
required outputs or including unnecessary
columns?

2) Constraint Fidelity
Are all filters, constraints, and qualifiers
expressed or implied in the question
faithfully reflected in the SQL logic?

3) Structural Alignment
Do joins, aggregations, groupings,
calculations, and subqueries match the
analytical intent of the question?

4) Unnecessary Complexity
Does the SQL introduce unnecessary joins,
filters, columns, subqueries, or computations
that are not required by the question and
could alter meaning or increase risk?

RATING SCALE
- Excellent: Fully correct, no issues
- Good: Minor issues, no material impact
- Average: Partial correctness with
  noticeable issues
- Poor: Major issues or mostly incorrect

OUTPUT INSTRUCTIONS
Return ONLY a valid JSON object.
Explanations must be objective and cite
specific aspects of the SQL or question.

Output JSON format ONLY:

{
  "Result Completeness": {
    "level": "Excellent | Good | Average | Poor",
    "explanation": "string"
  },
  "Constraint Fidelity": {
    "level": "Excellent | Good | Average | Poor",
    "explanation": "string"
  },
  "Structural Alignment": {
    "level": "Excellent | Good | Average | Poor",
    "explanation": "string"
  },
  "Unnecessary Complexity": {
    "level": "Excellent | Good | Average | Poor",
    "explanation": "string"
  }
}
\end{verbatim}
}

\subsection{LLM-as-a-Judge Prompt--Question Realism}
{\small
\begin{verbatim}
ROLE
You are a Senior Business Intelligence Lead
with extensive experience receiving analytics
requests from business stakeholders through
dashboards, BI tools, emails, and written
specifications.

Your task is to assess the BUSINESS REALISM
of a natural-language question.

You are given a business domain and a
question. Evaluate the question based on
professional business standards, NOT on
technical feasibility.

CONTEXT
BUSINESS DOMAIN:
{{DOMAIN_NAME}}

QUESTION:
{{question}}

EVALUATION DIMENSIONS

1) Business Relevance
Does the question reflect a real business
decision, concern, or workflow?

- Excellent:
  Clearly motivated by a real business need.
  A professional could reasonably ask this
  during daily work.
- Poor:
  Academic, purposeless, or purely exploratory.
  Lacks clear business motivation.

2) Business Language
Does the question use language, metrics,
and concepts natural to this domain?

- Excellent:
  Uses domain-appropriate terminology and
  KPIs correctly and naturally.
- Poor:
  Generic, awkward, schema-driven, or
  misuses business terms.

3) Natural Expression
Is the question phrased in a natural,
human-written manner?

- Excellent:
  Clear, professional, and natural.
  May be detailed if appropriate.
- Poor:
  Robotic, templated, or resembles SQL
  translated directly into English.

4) Decision Value
Would answering this question lead to
insight, prioritization, or follow-up action?

- Excellent:
  Clearly tied to a goal, KPI, or pain point,
  and supports decision-making.
- Poor:
  Overly broad, unbounded, or a data dump
  with limited decision value.

RATING SCALE
- Excellent:
  Indistinguishable from a real question
  asked by an experienced professional
- Good:
  Mostly realistic but slightly generic
- Average:
  Plausible but clearly synthetic
- Poor:
  Unnatural, academic, or obviously AI-generated

OUTPUT INSTRUCTIONS
Return ONLY a valid JSON object.
Explanations should be objective and refer
to specific aspects of the question.

Output JSON format ONLY:

{
  "Business Relevance": {
    "level": "Excellent | Good | Average | Poor",
    "explanation": "string"
  },
  "Business Language": {
    "level": "Excellent | Good | Average | Poor",
    "explanation": "string"
  },
  "Natural Expression": {
    "level": "Excellent | Good | Average | Poor",
    "explanation": "string"
  },
  "Decision Value": {
    "level": "Excellent | Good | Average | Poor",
    "explanation": "string"
  }
}

\end{verbatim}
}

\subsection{LLM-as-a-Judge Prompt--Question-Persona Alignment}
{\small
\begin{verbatim}
ROLE
You are a senior business analyst and
organizational workflow expert.

Your task is to evaluate whether a
natural-language question is well aligned
with a specific job role (persona) and one
of their real work scenarios.

You are given:
- A business domain
- A specific persona (job role)
- One work scenario for that persona
- The detailed workflow tasks
- A natural-language question

CONTEXT
DOMAIN:
{{DOMAIN_NAME}}

PERSONA:
{{persona_block}}

WORK SCENARIO:
{{scenario_block}}

WORKFLOW TASKS:
{{tasks_block}}

NATURAL-LANGUAGE QUESTION:
{{question}}

EVALUATION DIMENSIONS
Evaluate alignment between the question
and the persona/scenario along the
following aspects:

1) Role Responsibility
- Does the question fall within the typical
  decision scope and responsibilities of
  this role?
- Is it too detailed or too technical for
  this persona?

2) Scenario Relevance
- Does the question naturally arise in this
  work scenario?
- Would it make sense to ask while
  performing the listed tasks?

3) Practical Value
- Would the answer help the persona make
  a decision, take action, monitor progress,
  or assess outcomes in this scenario?

4) Role Language
- Is the question phrased in a way that
  matches how someone in this role would
  typically think and communicate?
- Avoid overly technical or database-
  oriented wording.

DO NOT EVALUATE
- Whether the question is answerable by
  a specific database
- Whether a SQL query exists or is correct

RATING SCALE
- Excellent:
  A very natural and expected fit for this
  role and scenario; routinely asked in
  practice
- Good:
  Generally appropriate but with minor
  issues in focus or phrasing
- Average:
  Related but mis-scoped, uncommon, or
  only marginally relevant
- Poor:
  Not appropriate for this role or scenario

OUTPUT INSTRUCTIONS
Return ONLY a valid JSON object.
Explanations should be objective and refer
to specific aspects of the question.

Output JSON format ONLY:

{
  "Role Responsibility": {
    "level": "Excellent | Good | Average | Poor",
    "explanation": "string"
  },
  "Scenario Relevance": {
    "level": "Excellent | Good | Average | Poor",
    "explanation": "string"
  },
  "Practical Value": {
    "level": "Excellent | Good | Average | Poor",
    "explanation": "string"
  },
  "Role Language": {
    "level": "Excellent | Good | Average | Poor",
    "explanation": "string"
  }
}

\end{verbatim}
}